\documentclass[10pt,twocolumn,letterpaper]{article}

\usepackage{cvpr}
\usepackage{times}
\usepackage{epsfig}
\usepackage{graphicx}
\usepackage{amsmath}
\usepackage{amssymb}
\usepackage[titletoc,title]{appendix}
\usepackage{comment}
\usepackage{array}
\usepackage{multirow}

\newcommand{\slamrnn}{GRNNs }
\newcommand{\slamrnnE}{GRNNs}

\newcommand{\depth}{D}

\newcommand{\VV}{\mathrm{V}}
\newcommand{\VVb}{\VV^\depth}

\newcommand{\VVtwo}{\mathrm{\bar{V}}}

\newcommand{\mem}{\mathrm{m}}
\newcommand{\softmax}{\mathrm{softmax}}

\usepackage[pagebackref=true,breaklinks=true,letterpaper=true,colorlinks,bookmarks=false]{hyperref}

\cvprfinalcopy 

\def\cvprPaperID{3877} 

\begin{document}

\title{Learning Spatial Common Sense with Geometry-Aware Recurrent Networks}

\author{ Hsiao-Yu Fish Tung\textsuperscript{1}\thanks{Indicates equal contribution}
\quad
Ricson Cheng \textsuperscript{2}\footnotemark[1]\,\,\thanks{Work done while at CMU}
\quad
Katerina Fragkiadaki \textsuperscript{1}\\
{ \textsuperscript{1}Carnegie Mellon University \quad 
\textsuperscript{2}Uber Advanced Technologies Group}\\
{\tt\small \{htung,katef\}@cs.cmu.edu, ricsonc@uber.com}\\
}

\maketitle

\begin{abstract}
    We integrate two powerful ideas, geometry and deep visual representation learning, into recurrent network architectures for mobile visual scene understanding.  
    The proposed networks learn to ``lift" and integrate 2D visual features over time into  latent 3D feature maps of the scene.  
    They 
    are equipped with differentiable geometric operations, such as projection, unprojection, egomotion estimation and stabilization, in order to compute a geometrically-consistent mapping between the world scene and their 3D latent feature state.  
We train the proposed architectures to predict novel camera views given short frame sequences as input. Their  predictions strongly generalize to scenes with a novel number of objects, appearances and configurations; they greatly outperform  previous works that do not consider egomotion stabilization or a space-aware latent feature state.   
 We train the  proposed architectures to detect and segment objects in 3D using  
  the latent 3D feature map as input---as opposed to per frame features.   
 The resulting object detections 
 persist over time: they continue to exist even  when an object gets occluded or leaves the field of view. 
 Our experiments suggest the proposed space-aware latent feature memory and  egomotion-stabilized convolutions are essential architectural choices for spatial common sense to emerge 
 in artificial embodied visual agents. 
 
\end{abstract}

\section{Introduction} \label{sec:intro}
 \begin{figure}[h!]
    \centering
    \includegraphics[width=0.48\textwidth]{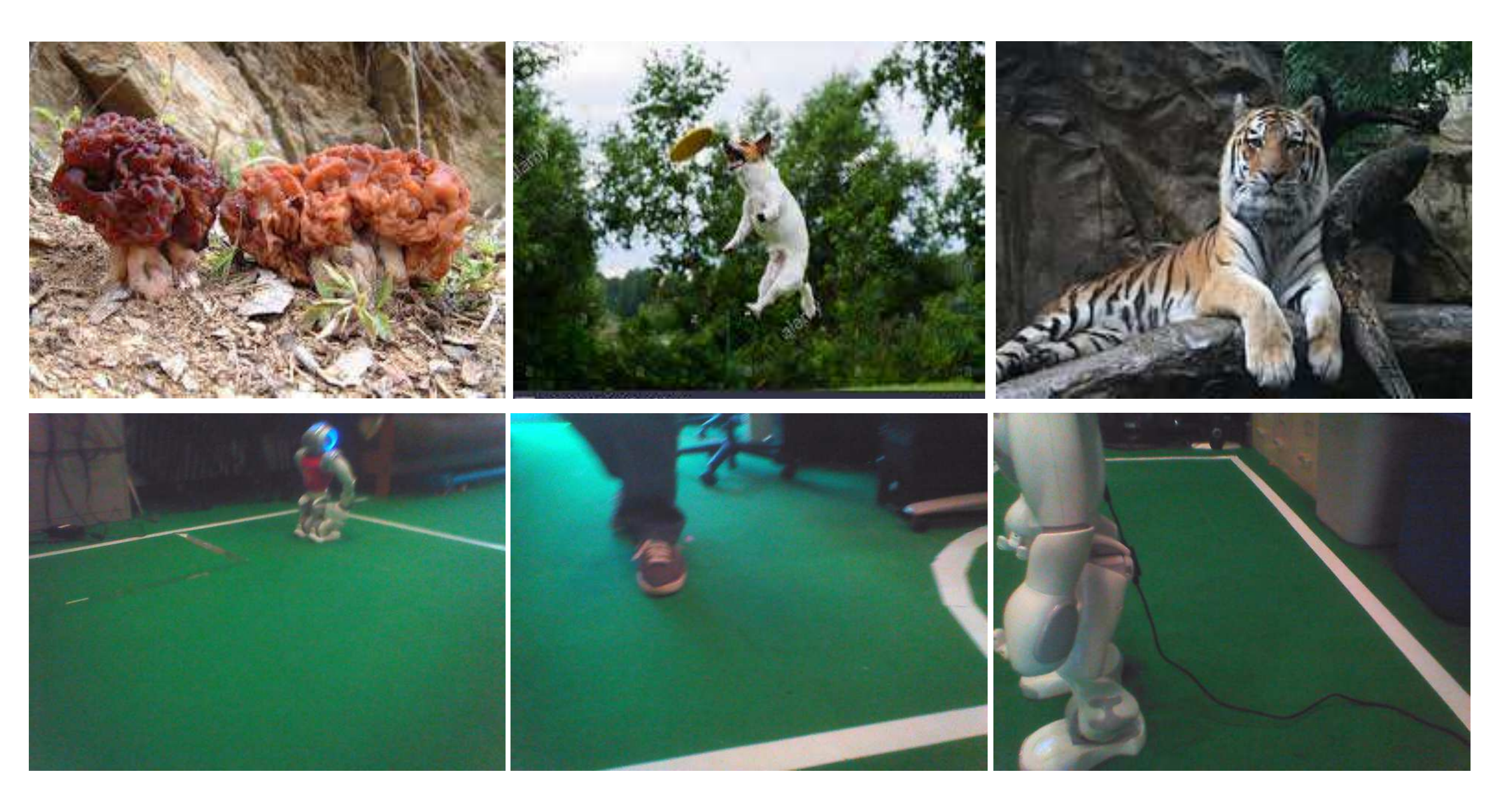}
      \caption{\textbf{Internet vision versus robotic vision}. 
     Pictures taken by humans  (top row)  (and uploaded on the web)  are the \textit{output} of   visual perception of a well-trained  agent, the human photographer. The content is skillfully framed  and the objects appear in canonical scales and poses. 
     Pictures taken by mobile agents, such as a NAO robot during a robot soccer game (bottom row), are the \textit{input} to such visual perception. The objects are often partially occluded and appear in a wide variety of locations, scales and poses.  
     We present recurrent neural architectures for the latter, that integrate visual information over time  to piece together the visual story of the scene. 
    }
    \label{fig:roboticvision}
\end{figure}

Current state-of-the-art visual systems \cite{DBLP:journals/corr/HeGDG17}  accurately detect object categories that are rare and unfamiliar to many of us,  such as \textit{gyromitra}, a particular genus of mushroom (Figure \ref{fig:roboticvision} top left). Yet, they neglect the basic principles of object permanence or  spatial awareness  that a one-year-old child has developed: once the camera turns away, or a person walks in front of the gyromitra, its detection disappears and it is replaced by the objects detected in the new visual frame. 
We believe the 
ability of current visual systems to detect  rare and exquisite object categories and their   inability to carry out elementary spatial reasoning 
 is due to  the fact that they  are trained 
 to {\it label object categories} from {\it static  Internet photos} (in  ImageNet and COCO datasets) using a {\it single frame} as input. 
Our overexposure to Internet photos  makes us  forget how pictures captured by mobile agents look. Consider Figure \ref{fig:roboticvision}. Internet  photos are skillfully captured by human photographers, are well framed and show objects unoccluded, in canonical locations, scales and poses (top row). Instead, photos captured by NAO robots during a soccer game  show objects in a wide variety of  scales, poses,  locations, and occlusion configurations (bottom row).  
Often, it would not even make sense to  label objects in such images, as most  objects appear only half-visible.  In the case of Internet vision, the picture is the \textbf{output} of visual perception of a well-trained visual agent, the human photographer. In the case of mobile robotic vision, the picture is the \textbf{input} to such visual perception.  Thus, different architectures may be needed for each.

\begin{figure}[t!]
    \centering
    \includegraphics[width=0.5\textwidth]{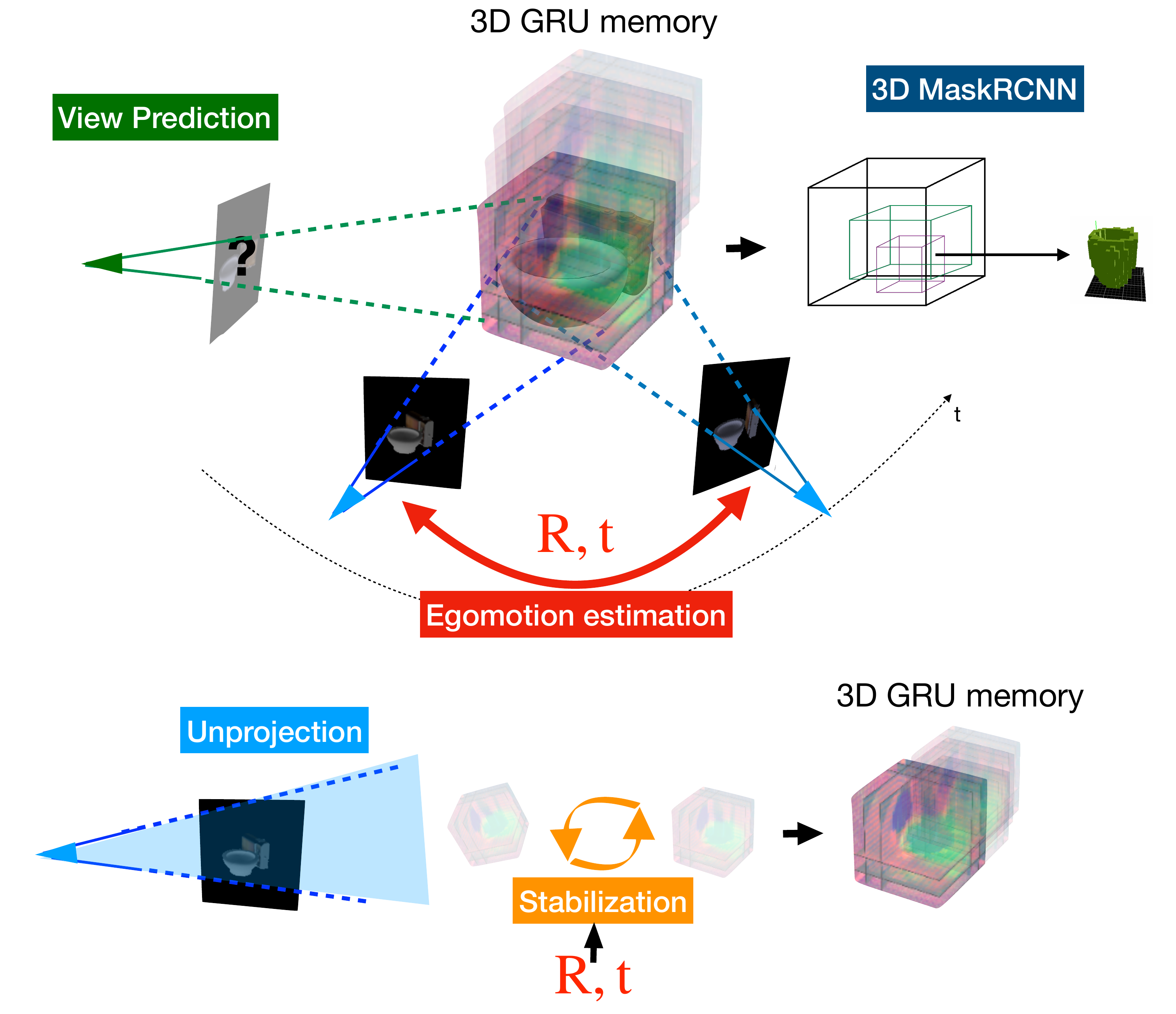}
    \caption{\textbf{Geometry-aware Recurrent Neural Networks (\slamrnnE)} integrate visual information over time in a 3D  geometrically-consistent deep feature  memory of the visual scene. At each frame, RGB images are \textit{unprojected} into  corresponding 3D feature tensors, which  are  oriented  to the coordinate frame of the memory map built thus far (2nd row). 
    A 3D convolutional GRU memory is then updated using the egomotion-stabilized features as input. 
    }
    \label{fig:overview}
\end{figure}

We present Geometry-aware Recurrent Neural Network architectures, which we call \slamrnnE, that learn  to ``lift" and integrate over time  2D image features into 3D feature maps of the scene, while stabilizing against the egomotion of the agent.  
They update over time a 3-dimensional  latent feature state: the latent feature vectors are arranged in a 3D grid, where every location of the grid encodes a 3D physical location in the scene. The latent state is updated with each new input frame using egomotion-stabilized convolutions, as shown in Figure \ref{fig:overview}.  
\slamrnn learn to  map   2D input visual features  to a 3D latent feature map, and back, in a differentiable manner.  
To achieve such differentiable and geometrically-consistent mapping between the world scene and the 3D latent feature state, they are equipped with differentiable geometric operations, such as egomotion estimation and feature stabilization, 3D-to-2D projection, and 2D-to-3D unprojection, as shown in Figure \ref{fig:overview}. Beyond being space-aware, we do not impose any other constraints on the learned representations: they are free to encode  whatever is  relevant for the downstream task.

We train \slamrnn in a self-supervised manner to  predict image views from novel camera viewpoints, given short frame sequences as inputs. We empirically show 
\slamrnn  learn to predict novel views and  \textbf{strongly generalize} to novel scenes with different  number, appearances and configuration of objects. They greatly outperform  geometry-unaware networks of previous works 
that are trained under the  exact same view-prediction loss, but do not use  egomotion-stabilized convolutions or a 3D latent space. We argue  strong generalization is a necessary condition for claiming the ability to spatially reason. 
Furthermore, we show  learnt representations of GRNNs support scene arithmetics: adding/subtracting latent scene 3D feature maps and decoding  them  from a particular viewpoint matches the result of adding/subtracting 3D world scenes directly.

We train \slamrnn in  a supervised manner  to detect and segment   objects in 3D,  given short frame sequences as inputs (Figure \ref{fig:overview}). We use the latent 3D feature map  as input to a 3D version of maskRCNN \cite{DBLP:journals/corr/HeGDG17}, a state-of-the-art 2D object detector/segmentor. 
The resulting 3D object detections and 3D voxel segmentations   \textit{persist in time} despite instantaneous occlusions and dis-occlusions: an object that is not visible in the current frame  is still present in the latent 3D feature map. By projecting the detected 3D objects in 2D  we obtain \textit{amodal} \cite{Regolin1995} object boxes and segments, even under severe occlusions.  Visual recognition with GRNNs thus exhibits  \textbf{object permanence}, a property which is effortless for  humans, and impossible thus far for 2D visual detectors. 
The GRNN architecture borrows many ideas from SLAM (Simultaneous Localization and Mapping) methods, as we discuss in Section \ref{sec:model}. GRNNs though  eventually learn to ``imagine'' missing content of the scene without  visual inspection from multiple viewpoints. Datasets and code are available at our project page  \url{https://ricsonc.github.io/grnn/}.

\section{Related  Work} \label{sec:related}

\paragraph{Deep geometry} 

Simultaneous Localization and Mapping (SLAM) \cite{schoeps14ismar,kerl13iros} methods  are purely geometric methods that build a 3D pointcloud map of the scene while estimating the motion of the camera. Our method builds multiple deep feature maps  instead, which capture both the geometry and the semantics of the scene. 
Recently, there has  been great interest in integrating learning and geometry for single view 3D object reconstruction \cite{DBLP:journals/corr/TulsianiZEM17, DBLP:journals/corr/abs-1711-03129}, 3D object reconstruction from videos \cite{DBLP:journals/corr/NovotnyLV17a}, depth and egomotion estimation from pairs of frames \cite{sfmnet,tinghuisfm}, depth estimation from stereo images \cite{DBLP:journals/corr/GodardAB16}, and estimation of 3D  human keypoints from 2D keypoint heatmaps \cite{3Dinterpreter,DBLP:journals/corr/TungHSF17}. Many of those works use neural network architectures equipped with some form of differentiable camera projection, so that the 3D desired estimates can be supervised directly using 2D quantities.  
For example, Tulsiani~\etal~\cite{DBLP:journals/corr/TulsianiZEM17}, Wu~\etal~\cite{ DBLP:journals/corr/abs-1711-03129} and Zhou~\etal\cite{tinghuisfm} use a single image frame as input to predict a 3D reconstruction for a single object, or a 2D depth map for the entire scene. 
These works use multiple views only to obtain extra regularization for the predictions in the form of depth re-projection error. 
Learnt stereo machines (LSM) \cite{LSM} integrate RGB information along sequences of random camera viewpoints into a latent  3D feature memory tensor, in an egomotion-stabilized way, similar to our method. However, their goal is to 3D reconstruct a single object, as opposed to detect and 3D reconstruct multiple objects, which our model does. They assume egomotion is given, while we also propose a way to estimate egomotion.  They can only be trained supervised for the object 3D reconstruction task, while \slamrnn can be trained self-supervisedly through view prediction. The work of LSM has inspired though the models proposed in this paper.




MapNet \cite{henriques2018mapnet}, Cognitive mapping and planning \cite{gup}, IQA \cite{DBLP:journals/corr/abs-1712-03316} and Neural Map \cite{DBLP:journals/corr/ParisottoS17} construct 2D overhead maps of the scene by taking into account the egomotion of the observer, similar to our method. MapNet further estimates the egomotion, while other methods assume it is  known. 
In IQA,  objects are detected in each  frame and  detections are aggregated in a  birdview map,  whereas we detect objects directly using the 3D feature map as input. 

The closest work to ours  is  the work of Cheng at al. \cite{activevision}, which considers egomotion-stabilized convolutions and a 3D latent map for segmenting objects in 3D, like us. However, they assume egomotion is known---while we learn to estimate it---and their object detection pipeline  uses  heuristics in order to specify  the number of objects in the scene by discretizing continuous voxel segmentation embeddings that they obtain with metric learning. We instead train 3D region proposal and segmentation networks. Most importantly, they do not consider self-supervised learning via view prediction, which is one of the central contributions of this work.  Rather, they exclusively focus on supervised voxel labelling using groundtruth 3D voxel occupancies provided by a simulator.

\paragraph{Self-supervised visual feature learning}
Researchers have considered many self-supervised tasks to train visual representations without human labels. 
For example, works of \cite{DBLP:journals/corr/JayaramanG15,DBLP:journals/corr/AgrawalCM15}  train visual representation by predicting egomotion between consecutive frames, and works of \cite{Eslami1204,DBLP:journals/corr/TatarchenkoDB15}  predict novel views of a scene. 
In particular, the authors of generative query network (GQN) \cite{Eslami1204} argue that GQN learns to disentangle color, lighting, shapes and spatial arrangement without any human labels. We compare against their model in Section \ref{sec:exp} and show \slamrnn can strongly generalize beyond the training set, while GQN cannot. Such strong generalization suggests that 3D latent space and egomotion-stabilization  are necessary architectural choices for spatial reasoning to emerge. 


\paragraph{3D object detection}
When LiDAR input is available, many recent works attempt detecting  objects directly in 3D  
 using LiDAR and RGB streams \cite{Zhou2017VoxelNetEL, Liang_2018_ECCV, pmlr-v87-yang18b}. 
They mostly use a single frame as input, while the proposed \slamrnn integrate visual information over time. 
Extending \slamrnn to  scenes with independently moving objects is a clear avenue for future work. 
\section{Geometry-aware recurrent networks} \label{sec:model}

\begin{figure*}[t!]
    \centering
    \includegraphics[width=1.0\textwidth]{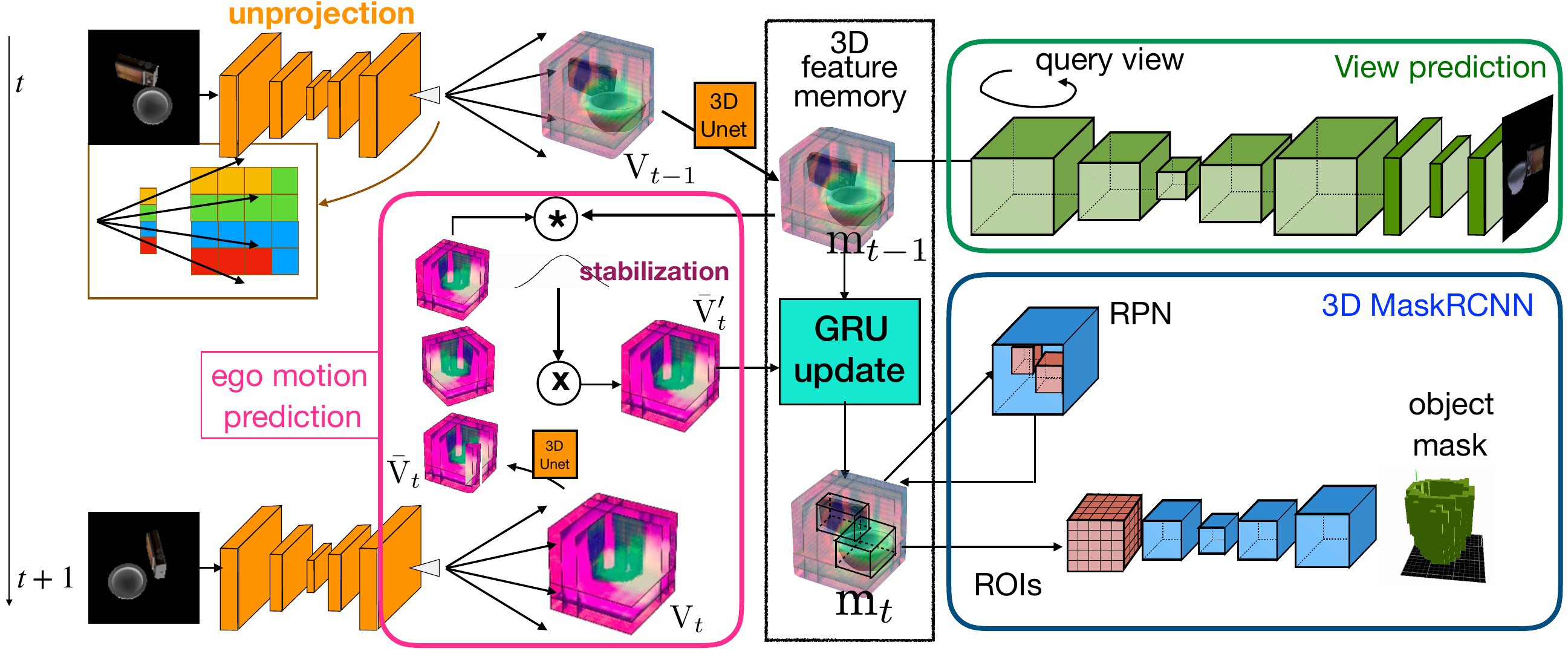}
    \caption{ \textbf{GRNN architecture.} 
    At each time step $t$, an RGB image $I_t$ is the input to a 2D U-net. The resulting 2D deep feature maps are unprojected to 4D tensors $\VV_t$, which in turn are input to a 3D U-net (we do not show the optional combination with unprojected depthmaps for clarity).  The resulting 3D deep feature maps $\VVtwo$ are oriented to cancel the relative camera motion between the current viewpoint and the coordinate system of the 3D GRU memory state $\mem_{t-1}$, as estimated by an egomotion estimation module. The resulting oriented 3D deep feature maps $\VVtwo_t$' update the 3D GRU memory state and output $\mem_{t}$. The updated state of the GRU module is then projected from specific viewpoints and decoded into a corresponding RGB image for view prediction, or fed into a 3D MaskRCNN to predict 3D object bounding boxes and object voxel occupancies.  }
    \label{fig:arch}
\end{figure*}

 \slamrnn are recurrent neural networks whose latent state $\mem_t \in \mathbb{R}^{w \times h \times d \times c}, t=1 \cdots T$  learns a 3D deep feature map of the visual scene. We use the terms 4D tensor and 3D feature map interchangeably, to denote a set of feature  channels, each being 3-dimensional. The memory map is updated with each new camera view in a geometrically-consistent manner, so that information from 2D pixel projections that correspond to the same 3D physical point end up nearby in the memory tensor, as illustrated in Figure \ref{fig:arch}. This permits later convolutional operations to have a correspondent input across frames, as opposed to it varying with the motion of the observer. We believe this is a key for generalization. 
The main  components of \slamrnn are illustrated in Figure \ref{fig:arch} and are detailed right below.

\paragraph{Unprojection}
At each timestep, we feed the input RGB image $I_t$ to a 2D convolutional encoder-decoder network with skip-connections  (2D U-net \cite{DBLP:journals/corr/RonnebergerFB15}) to obtain a set of 2D feature maps $ \mathrm{F}_t \in \mathbb{R}^{w \times h \times c}$.  
We then unproject all 
feature maps 
to create a  4D feature tensor $\VV^I_t \in \mathbb{R}^{w \times h \times d \times c}$ as follows: 
For each "cell" in the 3D feature grid indexed by $(i,j,k)$, we compute the 2D pixel location $(x,y)$ which the center of the cell projects onto, from the current camera viewpoint:
\begin{equation}
    [ x,y ]=  [f \cdot i/k, f \cdot j/k], \nonumber
\end{equation}
where $f$ is the focal length of the camera. 
Then, $\VV^I_{i,j,k,:}$ is filled with the bilinearly interpolated 2D feature vector at that pixel location $(x,y)$. All voxels lying along the same ray casted from the camera center will be filled with nearly the same image feature vectors. 
We further  unproject the input 2D depthmap $\depth_t$ 
into 
a binary voxel occupancy grid $\VVb_t \in \{0,1\}^{w \times h \times d}$ that contains the thin shell of voxels directly visible from the current camera view. We compute this by filling all voxels whose unprojected  depth value equals the grid depth value. When a depth sensor is not available, we learn to estimate the depthmap  using a 2D U-net that takes the RGB image as input. 

We multiply each 3-dimensional channel of the feature tensor $\VV^I_t$ with the binary occupancy grid $\VVb_t$ to get a final 4D   feature tensor $\VV_t \in \mathbb{R}^{w \times h \times d \times c}$.
The unprojected tensor $\VV_t$ enters a 3D encoder-decoder network with skip connections (3D U-net) to produce a resulting feature tensor $\VVtwo_t \in \mathbb{R}^{w \times h \times d \times c}.$

\paragraph{Egomotion estimation and stabilization}
Our model orients the 3D feature memory  to have 
$0^{\circ}$  elevation using the absolute elevation angle of the first camera view.  
We assume this value is given, but it can also be estimated using a 2D convnet. 
This essentially makes the memory  to always be  \textit{parallel to the ground plane}. The azimuth of the 3D feature memory is chosen to be the azimuth of the 
first view in the input frame sequence. We assume  the camera does not translate, only rotates by varying two degrees of freedom, elevation and azimuth. 

At each time step $t$, we  estimate the relative elevation and azimuth between the current frame's viewpoint and the feature memory. 
Note that we can alternatively predict the (absolute) elevation directly from each input view, without matching against the memory built thus far. For the azimuth, since we need to estimate  the relative azimuth to the first view, such cross-view comparison is necessary. 
Specifically, the tensor $\VVtwo_t$ is \textit{rotated} by different azimuth and elevation angles  and results in a stack of rotated feature tensors $\VVtwo^{\mathrm{rot}} \in \mathbb{R}^{(L \cdot K)\times w \times h \times d \times c}$, where $L,K$ are the total number of azimuths and elevation angles considered, respectively, after discretization. Similar to the bilinear interpolation used during  unprojection, to fill in each feature voxel in a rotated tensor $\VVtwo^{\mathrm{rot}}_{\cdot,i,j,k,:}$, we compute the 3D location $(X,Y,Z)$ where it is rotated from and insert the bilinearly interpolated feature value from the original tensor $\VVtwo_t$. 
We then compare each of the rotated feature maps with our current 3D feature memory $\mem_{t-1} \in \mathbb{R}^{w \times h \times d \times c}$ using matrix inner products, to produce a probability distribution over azimuth and elevation pairs:
\begin{align}
    \bar{\rho}_t(r) &= \mem_{t-1} * \VVtwo_{rot}(r,:,:,:,:), \quad r \in 1 \cdots L \cdot K \nonumber \\
    \rho_t &= \softmax(\bar{\rho}_t) \nonumber,
\end{align}
where $*$ denotes matrix inner product. The resulting  rotation  $\bar{r}_t$ is obtained by a weighted average of azimuth and elevation  angles where weights are in $\rho_t$.  
Finally, we orient the  tensor $\VVtwo_t$ to cancel the relative rotation $\bar{r}_t$ with respect to our 3D memory  $\mem_{t-1}$, we denote the oriented tensor as   $\VVtwo'_t$.

\paragraph{Recurrent map update}

Once the feature tensor has been properly oriented, 
we feed $\VVtwo'_t$ as input to a 3D convolutional Gated Recurrent Unit  \cite{DBLP:journals/corr/ChoMGBSB14} layer, whose hidden state is the memory  $\mem_{t-1} \in \mathbb{R}^{w \times h \times d \times c}$, as shown in Figure \ref{fig:arch}. This state update outputs $\mem_{t}$.   The hidden state is initialized to zero at the beginning of the frame sequence. 
For our view prediction experiments (Section \ref{sec:exp}) where we use a fixed number of views $T$, we found that averaging, namely 
$\mem_T = \frac{1}{T} \sum_t \bar \VVtwo'_t$
works equally well to using the GRU update equations, while being much faster. 

\paragraph{Projection and decoding}
Given a 3D feature memory $\mem_t$ and a desired viewpoint $q$, we first rotate the 3D feature memory so that its depth axis  is aligned with the query camera axis. We then generate for each depth value $k$ a corresponding  projected feature map $\mathrm{p}_k \in \mathbb{R}^{w \times h  \times c}$. Specifically, for each depth value, the projected feature vector at a pixel location $(x, y)$ is computed by first obtaining the 3D location it is projected from 
and then inserting bilinearly interpolated value from the corresponding slice of the 4D tensor $\mem$. In this way, we obtain $d$ different projected maps, each of dimension $w \times h \times c$. Depth ranges from $D-1$ to $D+1$, where $D$ is the distance to the center of the feature map, and are equally spaced.

Note that we do not attempt to determine visibility of features at this projection stage. The stack of projected maps is processed by 2D convolutional operations and is decoded using a residual convLSTM decoder, similar to the one proposed in \cite{Eslami1204}, to an RGB image. We do not supervise visibility directly. The  network implicitly learns to determine visibility and to choose appropriate depth slices from the stack of projected feature maps. 

\subsection{View prediction}

Mobile agents have access to their egomotion, and can observe  sensory outcomes of their motions and interactions.  
Training sensory representations to predict such outcomes
is a  useful form of supervision,  free of human annotations,  often termed  \textit{self-supervision} since the ``labels" are provided by the embodied agent herself.  Can spatial common sense, the notion of objects and scenes, geometry, visibility and occlusion relationships,  
emerge in a self-supervised way in a mobile agent that moves around and observes the world? 

We train  \slamrnn to predict the image the agent would see from a novel viewpoint, given a short view sequence as input.  Given the  3D feature memory  and  a query viewpoint, we orient the map to the query viewpoint, we project it to 2D and decode it to an RGB image,  as described above. We train our view prediction using a standard cross-entropy pixel matching loss, where the pixel intensity has been squashed  into the range $[0,1]$. Our model is end-to-end differentiable. For view prediction, we did not use depth as input, nor did we use a 2D U-net to estimate it. We also did not set the memory to be parallel top the ground plane.  We use only the RGB input and set the coordinate system of the memory to match that of the first camera view after unprojection, for a fair comparison with prior art. We show in Section \ref{sec:exp} that  \slamrnn  greatly outperform alternative geometry-unaware RNN architectures in view prediction and \textbf{strongly generalize} beyond the training set to novel scenes with different number of objects, appearances and arrangements. Training and implementation details are included in the supplementary file.

\subsection{3D object detection and segmentation}
We train \slamrnn  in a supervised manner to predict 3D object bounding boxes and 3D object segmentation masks, using groundtruth 3D object boxes and 3D voxel segmentations from a simulator. 
We adapt MaskRCNN \cite{DBLP:journals/corr/HeGDG17}, a state-of-the-art  object detector/segmentor,  to have 3D input and output, instead of 2D. 
Specifically, we consider every grid location $(X,Y,Z)$ in our 3D memory to be a candidate 3D box centroid.  
At each time step, the 3D feature memory  $\mem_t$ 
is fed to a 3D region proposal network to predict positive anchor centroids, as well as  the  
corresponding adjustment for the box center location  and the box  dimensions, width, height and depth. Our 3D bounding box encoding is similar to the one proposed in VoxelNet \cite{Zhou2017VoxelNetEL}. We filter the proposed boxes using non-max suppression to reject highly overlapping ones. We train with a combination of classification  and regression loss, following well established  detector training schemes \cite{DBLP:journals/corr/RenHG015,DBLP:journals/corr/HeGDG17}. 
The proposed 3D bounding boxes that have Intersection of Union (IoU) above a specific threshold with a corresponding groundtruth object  box are denoted as  Regions of Interest (ROIs) and are used to pool  features from their interior to predict  3D object voxel occupancy, as well as a second refinement of the predicted 3D box location and dimensions.

\paragraph{Object permanence} Even when an object is not visible in the current camera viewpoint, its features are  present in the 3D feature memory,  and our detector detects and segments it, as we show in the second column of Figure \ref{fig:bbox_visual}. In other words, object detections persist through occlusions and changes of the field of view caused by camera motion.  
Applying the detector on the latent 3D model
of the scene as opposed to the 2D visual frame is beneficial.
The latent 3D model follows the physical laws of 3D non-intersection and object permanence, while 2D visual observations do not.

\section{Experiments}\label{sec:exp}
\vspace{-0.1in}


  The term ``spatial common sense" is broad and concerns the ability to perceive and understand properties and regularities regarding spatial arrangements and motion that are shared by (``common to") nearly all people. Such common sense includes the fact that objects have 3D shape as opposed to being floating 2D surfaces, the fact that scenes are comprised of objects, the 3D non-intersection principle, the fact that objects do not spontaneously disappear, and many others   \cite{DBLP:journals/corr/abs-1212-4799}.
The   model we propose in this work targets understanding of \textit{static} scenes,  that is, scenes that do not contain  any independently moving objects, and that are  viewed under a potentially moving observer.  Thus, we restrict the term \textit{spatial common sense} to refer to rules and regularities that can be perceived  in static worlds.
Our experiments  aim to answer the following questions:
\begin{enumerate}
    \item Do \slamrnn learn spatial common sense?
    \item Are geometric structural biases necessary for spatial common sense to emerge?
    \item How well do \slamrnn perform on egomotion estimation and 3D object detection?
\end{enumerate}


\subsection{View prediction} 

We consider the following simulation datasets: 

\noindent
i) \textit{ShapeNet arrangement}  from ~\cite{activevision} that contains scenes with  synthetic 3D object models from ShapeNet~\cite{shapenet2015} arranged on a table surface.   
The objects in this dataset belong to four object categories, namely, cups, bowls, helmets and  cameras. We follow the same train/test split of ShapeNet~\cite{shapenet2015}  so that object instances which appear in the training  scenes do not appear in the test  scenes. Each scene contains two objects, and each image is rendered from a viewing sphere which has $3\times18$ possible views with 3 camera elevations $(20^{\circ}, 40^{\circ}, 60^{\circ})$ and 18 azimuths $(0^{\circ}, 20^{\circ}, \dots, 340^{\circ})$. There are 300 different scenes in the training set and 32 scenes
with novel objects in the test set.  

\noindent
ii) \textit{Shepard-metzler} shapes dataset from ~\cite{Eslami1204} that contains  scenes with seven colored cubes stuck together in random arrangements. We use the train and test split of ~\cite{Eslami1204}. 

\noindent
iii) \textit{Rooms-ring-camera} dataset from ~\cite{Eslami1204} that contains rooms  with  random floor and wall colors, in which there are variable numbers of objects with different shapes and colors.

We compare  \slamrnn  against the recent "tower" architecture of  Eslami et al. \cite{Eslami1204}, a 2D network  trained under a similar view prediction loss. At each time step, the tower architecture takes as input a 2D RGB image and performs a series of convolutions on it. The camera pose from which the image was taken is tiled along the width and height axes and then concatenated with the feature map after the third convolution. Finally, the feature maps from all views are combined via average pooling. 
Both our model and the baseline use the same autoregressive decoder network.   
For fairness of comparison, we use groundtruth egomotion rather than estimated egomotion in all view prediction experiments, and only RGB input (no depth input of depth estimation) for both our model and the tower baseline. 
In both the baseline and our model, we did not use any stochastic units for simplicity and speed of training. Adding stochastic units in both is part of our future work.

\begin{figure}[t!]
	\centering
	\includegraphics[width=0.48\textwidth]{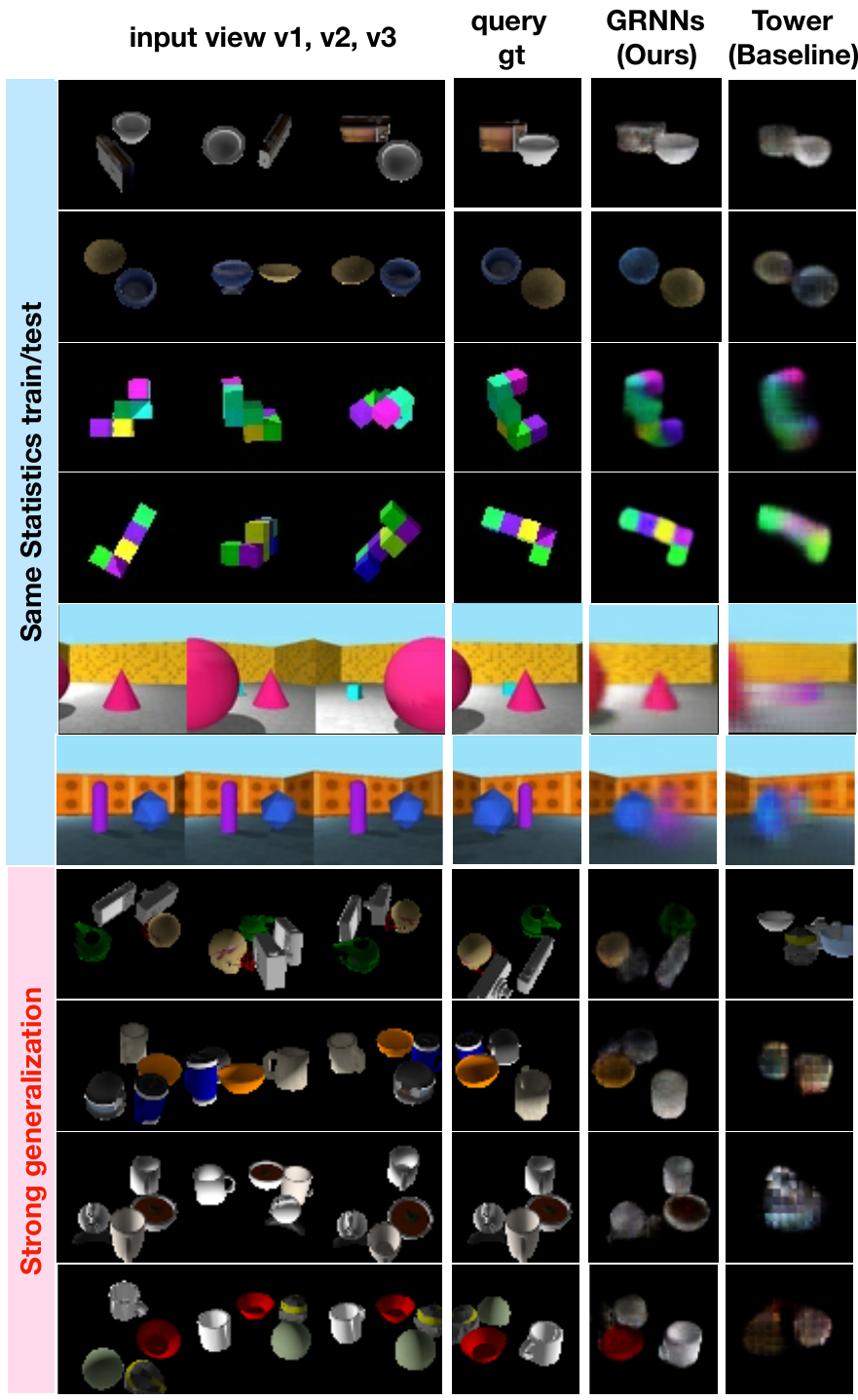}
\caption{
	\textbf{View prediction results} for the proposed \slamrnn and the tower model of Eslami et al. \cite{Eslami1204}. Columns from left to right show the three input views,  the groundtruth image from the query viewpoint,  the view predictions for \slamrnn and for the tower baseline. 
		The first two rows are from the  ShapeNet arrangement test set of \cite{activevision}, the next two rows are from the Shepard-Metzler test set of \cite{Eslami1204}, and the following two rows are from the \textit{Rooms-ring-camera} dataset also from ~\cite{Eslami1204}. The last four rows show generalization to  scenes with four objects from the  ShapeNet arrangement dataset, while both models were  trained only on scenes with two objects. \slamrnn outperform the baseline by a large margin and  \textit{strongly} generalize under a varying number of objects. 
	}
	\label{fig:viewprediction}
\end{figure}

Test results from our model and baseline on test images of ShapeNet arrangements and Shepard-metzler datasets are shown in Figure \ref{fig:viewprediction}.  Reconstruction test error for the ShapeNet arrangement test set is shown in Table \ref{tab:viewpredtable}. \slamrnn have a much lower reconstruction test error than the tower baseline. In Figure \ref{fig:viewprediction}, in the first four rows,  the distribution of the test scenes matches the training scene  distribution. Our model  outperforms the baseline in visual fidelity.  
In Figure \ref{fig:viewprediction}, in the last four rows, the test scene distribution does not match the training one: we test our model and baseline on  \textbf{scenes with four objects, while both models are trained on scenes with exactly two objects. } 
In this case, our model shows \textit{strong generalization} and outperforms by a margin the geometry-unaware baseline of  \cite{Eslami1204}, the latter refuses to see more than two objects present.  We argue the ability to spatially reason should not be affected by the number of  objects present in the scene. 
Our results  suggest that geometry-unaware models may be merely memorizing views with small interpolation capabilities, as opposed to learning to spatially reason.

\begin{figure}[h!]
	\centering
	\includegraphics[width=0.48\textwidth]{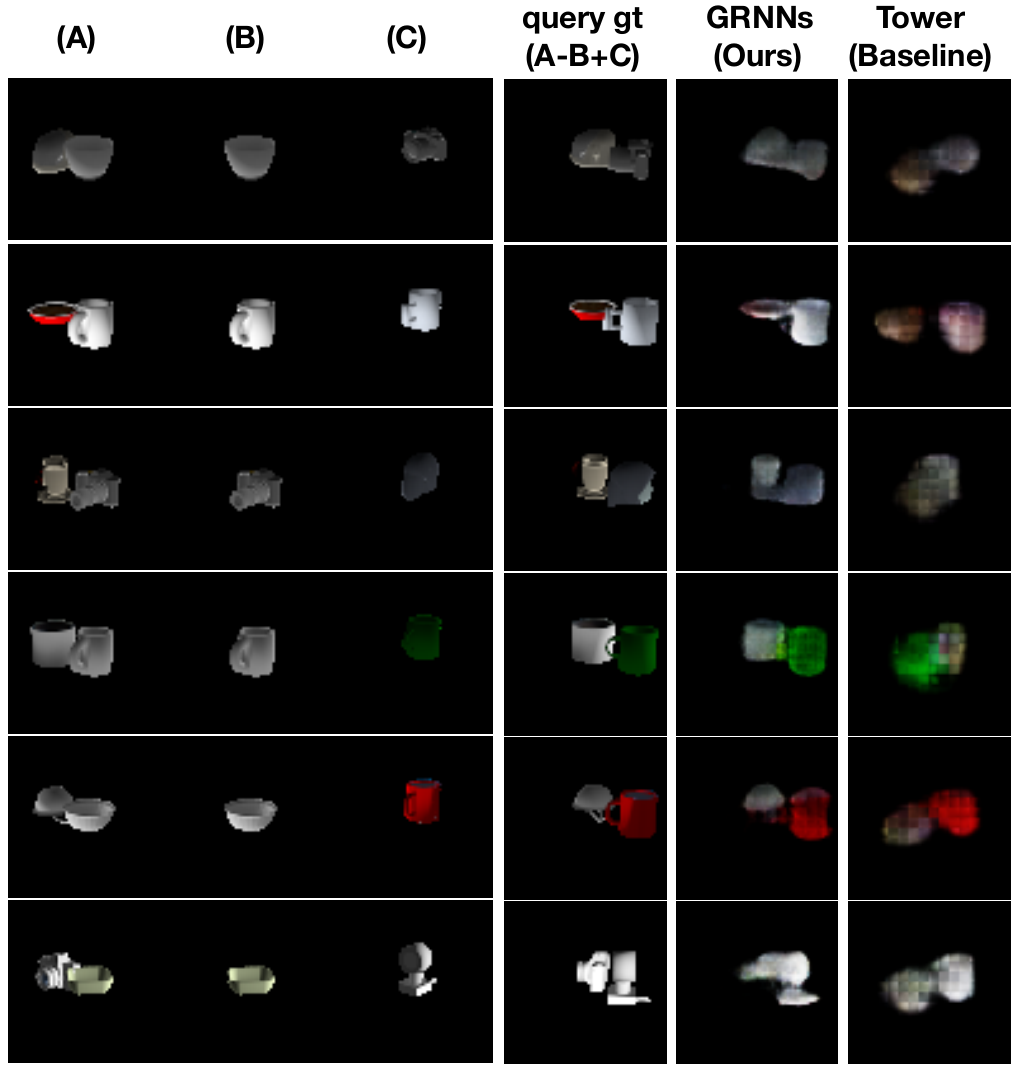}
	\caption{\textbf{Scene arithmetic} with \slamrnn and the model  of Eslami et al. \cite{Eslami1204} (tower). Each row is a separate "equation". We start with the representation of the  scene in the leftmost column, then subtract (the representation of) the scene in the second column, and add the (representation of the) scene in the third column. We decode the resulting  representation into an image. The groundtruth image is shown in the forth column. It is much more visually similar to the prediction of \slamrnn than to the tower baseline.}
	\label{fig:arithmetics}\label{key}
\end{figure}

\paragraph{Scene arithmetics}
The learnt  representations of \slamrnn  are capable of scene arithmetics, as we show in Figure \ref{fig:arithmetics}.  The ability to add and subtract individual objects from 3D scenes just by adding and subtracting their corresponding latent representations demonstrates that our model disentangles what from where. In other words, our model learns to store object-specific information in the regions of the  memory which correspond to the  spatial location of the corresponding object in the scene.  Implementation details and more qualitative view prediction results are included in the supplementary file.

\begin{table}[htbp!]
\centering
\begin{tabular}{c|cc}
            \multicolumn{1}{l|}{}    
            &  \multicolumn{1}{l}{Tower} & \multicolumn{1}{l}{GRNNs} \\ 
            \multicolumn{1}{l|}{}    
            &  \multicolumn{1}{l}{ (Baseline)} & \multicolumn{1}{l}{ (Ours)} \\ \hline
ShapeNet & $0.109 \pm 0.029$ & ${\bf 0.084} \pm 0.017$ \\
Shepard-Metzler & $0.081 \pm 0.017$ & ${\bf 0.073} \pm 0.014$ \\
\end{tabular}
	\vspace{1mm}
\caption{\textbf{View prediction  loss and the standard deviation} for the ShapeNet arrangement test set for two-object test scenes. Our model and baseline were trained on  scenes that also contain two objects with different object instances.}
\label{tab:viewpredtable}
\end{table}

\begin{figure*}[t!]
    \centering
    \includegraphics[width=1.0\textwidth]{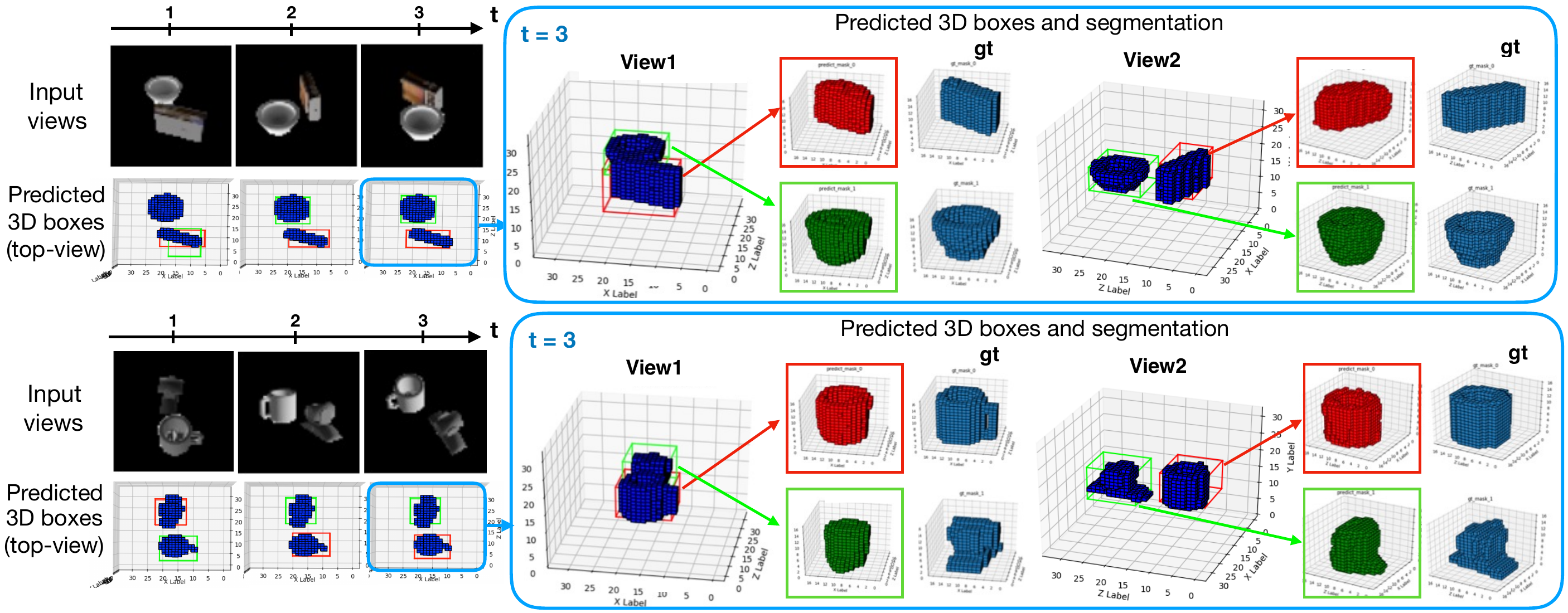}
     \caption{\textbf{3D object detection and segmentation} with GRNNs. In the first and  second row on the left we show the input images over time, and their corresponding object detection results for a top view, respectively. Blue voxels denote groundtruth objects and the predicted bounding boxes are shown in \color{red}red \color{black} and \color{green} green \color{black}.
     On the right, we show segmentation results for the third time step, visualizing the results from two views. Predicted 3D boxes and their corresponding predicted masks are show in red and green, and we show in blue the corresponding groundtruth. Best seen in color.
     }    
    
    \label{fig:bbox_visual}
    
\end{figure*}
 
\subsection{Egomotion estimation}
In this section, we quantify the error of our egomotion estimation component. We train our egomotion estimation module using groundtruth egomotion from a simulator, using the ShapeNet arrangement dataset. 
In Table \ref{tab:egotable}, we show egomotion estimation error in  elevation and azimuth angles. Our model improves its egomotion estimates with more views, since then a more complete feature memory is compared against each input unprojected tensor.  

\begin{table}[t!]
	\centering
	\begin{tabular}{c|ccc|c}
	\# views  & one & two & three & avg. \\
	 \hline
	 \slamrnn &
		 $8.6/ 17.8$ &
		 $5.6/16.8$ &
		 $5.6/ 6.6$ &
	 $6.6/13.7$ \\ \hline
	\end{tabular}
	\vspace{1mm}
\caption{ \textbf{Egomotion estimation error} of \slamrnn in elevation and azimuth angles for
the ShapeNet arrangement test set using different number of views. The error decreases with more views integrated in the memory.  }
	\label{tab:egotable}
\end{table}

\subsection{3D object detection and segmentation}

We use  the ShapeNet arrangement dataset, and the train/test scene split of \cite{activevision}.  We use mean Average Precision ($\mathrm{mAP}$) to score the performance of our model and baselines for  3D object  detection and 3D segmentation. Mean average precision measures the area under the precision-recall curve.
We vary the cutoff threshold of Intersection over Union (IoU) to be 0.33, 0.5 and 0.75 between our predictions and the groundtruth 3D boxes and masks.  We consider  four ablations for our model:  predicted egomotion (\textit{pego})  versus groundtruth egomotion (\textit{gtego}) used, and  predicted depth (\textit{pd}) versus groundtruth depth (\textit{gtd})  used as input. We use suffixes to indicate the model we use. 


\begin{table*}[htbp]
\centering

\begin{tabular}{|b{1.2cm}|p{1.3cm}p{1.3cm}p{1.3cm}p{1.3cm}||p{1.7cm}|p{1.3cm}p{1.3cm}p{1.3cm}p{1.3cm}|}
\hline
            detection           &2DRNN-gtego-gtd & GRNN-gtego-pd  & GRNN-gtego-gtd & GRNN-pego-gtd &
                      segmentation         & 2DRNN-gtego-gtd & GRNN-gtego-pd &GRNN-gtego-gtd &
                      GRNN-pego-gtd
                      \\
               
                      \hline
$\mathrm{mAP}_{0.75}^{d}$ & $0.364$ &  $0.471$ & ${\bf 0.816}$ & $0.549$ & $\mathrm{mAP}_{0.75}^{m}$ & $0.003$ &  $0.024$ &  ${\bf 0.058}$ & $0.023$ \\

$\mathrm{mAP}_{0.50}^{d}$ & $0.964$ &  $0.964$ & ${\bf 0.998}$ & $0.983$ &
$\mathrm{mAP}_{0.50}^{m}$ & $0.104$ &  $0.246$ & ${\bf 0.338}$ &0.249
\\
$\mathrm{mAP}_{0.33}^{d}$ & $0.998$ &  $0.994$ & ${\bf 0.999}$ & $0.999$ & 
$\mathrm{mAP}_{0.33}^{m}$ & $0.244$ &  $0.429$ & ${\bf 0.485}$ &0.384
\\ \hline
\end{tabular}
	\vspace{1mm}
\caption{Mean Average Precision (mAP) for 3D object detection and 3D segmentation for three different thresholds of Intersection over Union (IoU) (0.75,0.5,0.33) on ShapeNet arrangement test set of \cite{activevision}.}
\label{tab:maptable}
\end{table*}

 We compare against the following 2D baseline model, which we call \textit{2D-RNN}: we remove the unprojection, egomotion estimation and stabilization and projection operations from our model. The baseline takes as input an image and the corresponding depth map, feeds it to a 2D encoder-decoder network with skip connections to obtain a 2D feature tensor. The camera parameters for this view are concatenated as additional channels to the 2D feature tensor and altogether they are fed  to another 2D encoder-decoder network to obtain the 2D feature tensor for a 2D GRU memory update. 
We then feed the 2D memory feature tensor to an additional 2D encoder-decoder 
network and  reshape the channel dimension of its output into $d$ feature vector of length $7$ (one value for the anchor box prediction, six values for the 3D bounding boxes adjustments) to form a 4D tensor of size $w \times h \times d \times 7$ as prediction.

 We show mean average precision  for 3D object detection and 3D  segmentation for our model and the baseline in 
 Table \ref{tab:maptable},  and visualize  
 predicted 3D bounding boxes and segmentations from GRNNs (GRNN-gtego-gtd) in Figure \ref{fig:bbox_visual}. 
  GRNNs  significantly outperform the 2D-RNN.  Groundtruth depth input significantly helps  3D segmentation. This suggests that inferring depth using a cost volume as in \cite{LSM} would potentially help depth inference as opposed to relying on a per frame depthnet \cite{NIPS2014_5539} that does not have access to multiple views to improve its predictions.  
  Implementation details and more qualitative results are included in the supplementary file.

\section{Conclusion} \label{sec:conclusion}
\vspace{-0.1in}
We presented \slamrnnE, recurrent neural networks  equipped with differentiable geometric operations to estimate egomotion  and build 3D  deep feature maps for visual scene understanding on mobile visual agents. GRNNs add a new dimension to the latent space of previous recurrent models 
and ensure a geometrically-consistent mapping between the latent state and the 3D world scene. 
We showed  spatial common sense emerges in \slamrnn when trained in a self-supervised manner for novel view prediction. They can predict object arrangements, visibility and occlusion relationships in scenes with novel number, appearance and configuration of objects. We also showed that  view prediction as a loss does not suffice for spatial common sense to emerge, since 2-dimensional models of previous works fail to strongly generalize.  

Thus far, GRNNs  has been trained and tested on simulated scenes. Deploying our model on more realistic environments
is a clear avenue for future work.
We  expect pretraining in simulated environments to help performance in the real world. 
Besides, one limitation of the current model is that it operates on static scenes. 
Extending the proposed architectures to dynamic scenes, scenes with independently moving objects in addition to camera motion, 
is another very useful direction of future work. 
Finally, 
exploiting the sparsity of our  4D tensors to save GPU memory is an important direction for scaling up our model to  large scenes.

 \slamrnn  pave the way for embodied agents that learn visual representations and mental models by observing and moving in the world: these agents learn autonomously and develop the reasoning capabilities of young toddlers as opposed to merely mapping pixels to labels using human supervision.

\section*{Acknowledgement}
We would like to thank Xian Zhou for his help on training and testing the 3D MaskRCNN. This work is partly funded by a Google faculty award.

{\small
\bibliographystyle{ieee}
\bibliography{egbib}
}
\clearpage
{\Large \bf \centering \noindent Supplementary Material}

\appendix

\section{\slamrnn implementation details}

The input images, output images, and predictions (for view prediction) all have size 64 $\times $ 64. Our pre-unprojection 2D encoder-decoder network has encoder layers with 32, 64, 128, and 256 channels, respectively. The decoder layers are symmetric to the encoder layers. The sizes of these feature maps are 32 $\times $ 32, 16 $\times $ 16, 8 $\times $ 8 and 4 $\times $4 respectively, since each convolution has stride 2. For depth prediction, which is used for object detection, we use
the same 2D encoder-decoder network as described in this pre-unprojection step.

\paragraph{View prediction} We feed our feature tensor through a 3D encoder-decoder with skip-connections where the encoder has 64, 128, 256, 512, and 1024 channels respectively. The decoder is symmetric to the encoder. 
In our convLSTM decoder, we do not share weights across time steps.  Indeed, allowing each step to have different weights was noted by   ~\cite{Eslami1204} to improve performance. We use six generation steps, and we did not see noticeable improvement on image reconstruction quality  when more steps were used. We used 256 channels at each step. Both our model and the baseline are deterministic networks, we did not include any stochastic units in any of the two to accelerate training. 
We trained each model for 24 hours, using a batch size of 8 and a learning rate of $10^{-4}$. 
This resulted in roughly $800 \cdot 10^3$ steps of backpropagation for the tower (baseline) architecture and $160 \cdot 10^3$ steps for the GRNN architecture. We used the Adam optimizer.

\paragraph{Object detection/segmentation}
For object detection and ego-motion prediction, 
we feed the unprojected feature tensor 
through a 3D encoder-decoder with skip-connections where the encoder has 
16, 32, 64, 128 channels receptively, and the decoder is again symmetric to the encoder.
For ego-motion prediction, when we compare the current memory with 
the rotated feature tensors generated from the new view,
we use outputs from all the layers in the decoder (feature tensors with size 4 $\times$ 4 $\times$ 4 $\times$ 128,  8 $\times$ 8 $\times$ 8 $\times$ 64,  16 $\times$ 16 $\times$ 16 $\times$ 32, 32 $\times$ 32 $\times$ 32 $\times$ 16) to compute cross-convolutions.  
For object detection, we use only the last feature tensor from the 3D 
encoder-decoder as input and pass it to another 3D encoder-decoder with skip-connections to predict positive anchor centroids and their corresponding adjustments for the box centers. The channels in the second 3D encoder-decoder are set to 16, 32, 64, 128 and the corresponding final output is a  32 $\times$ 32 $\times$ 32 $\times$ 16 feature tensor. We then pass this
feature tensor to one more 3D
convolutional layer with kernel size 3, stride 1 and channel size 7.
The final prediction is a 32 $\times$ 32 $\times$ 32 $\times$ 7 
tensor with the first channel indicating positive anchor centroids and the last 6 channels indicating 3D boxes adjustments at each centroid. 
The model is trained with Adam optimizer (learning rate is set to $10^{-4}$)
without further parameter tuning. We train the model for roughly 25K iterations with batch size 2. We first train for  egomotion prediction, and next jointly train for egomotion prediction and object detection and segmentation losses. 

\section{Additional results}

\begin{figure*}[h!]
	\centering
	\includegraphics[width=0.83\textwidth]{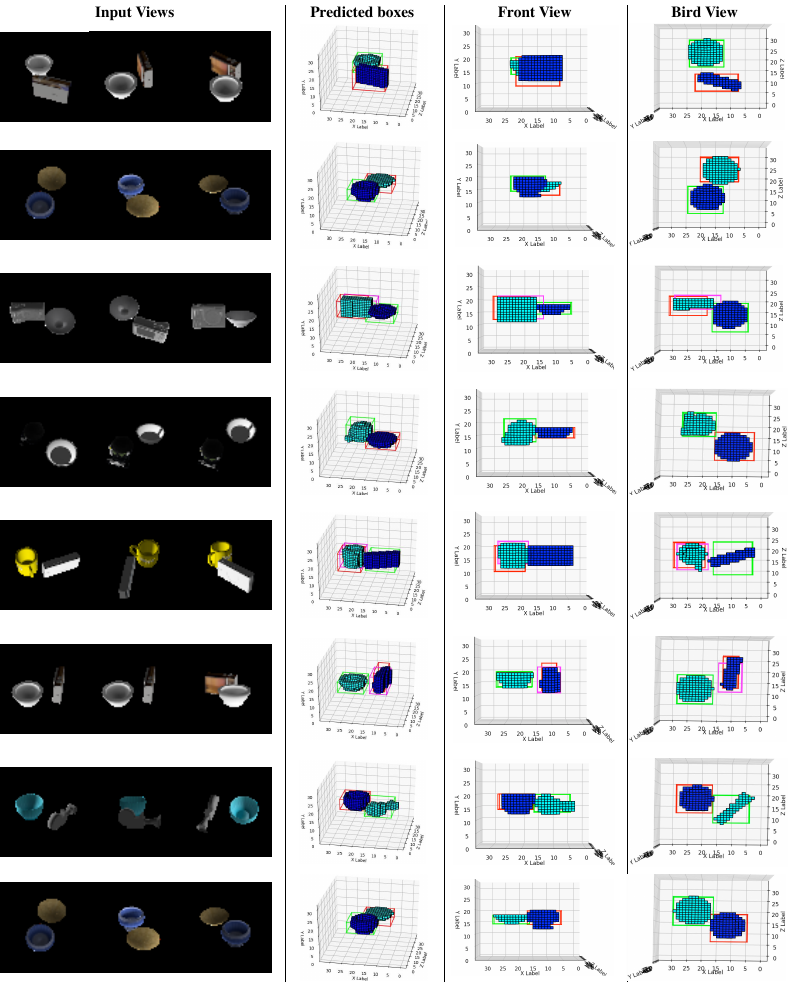}
	\caption{\textbf{Object detection and segmentation results.} Blue and light blue grids in the last three columns show {\bf groundtruth} voxel occupancy for the two objects present in the scene.  3D bounding boxes with different colors (red, green and magenta) are {\bf predicted} from the proposed 3D MaskRCNN. }
	\label{fig:detection}
\end{figure*}

In Figures ~\ref{fig:room_results}, \ref{fig:metzler_results} we show view prediction results on the \texttt{rooms\textunderscore ring\textunderscore camera} and \texttt{shepard-metlzer} dataset introduced in \cite{Eslami1204}. Since the camera intrinsics were not given, we used an estimated vertical and horizontal field of view of 60 degrees. Our model outperforms the baseline by a margin. In Figures  \ref{fig:shapnet_results}, \ref{fig:general_results}, we show more view prediction results on the  ShapeNet arrangement dataset of \cite{activevision}. In Figure \ref{fig:detection}, we show qualitative 3D object detection and segmentation results.

In Figure \ref{fig:singleinput}
we show, by using a {\it single} frame as input to our model, our model can generate completely novel views: our model can extrapolate the missing parts of the objects, which geometric SLAM methods \cite{xiang2017darnn, fusion++} cannot do. Our model not only remembers things it has seen in previous frames, but also learns to predict {\it invisible} parts of the scene. 

\newcommand{\image}[1]{
\begin{minipage}{.09\textwidth}
\includegraphics[width=\linewidth]{#1}
\end{minipage}
}

\newcommand{\generaterowsmall}[2]{
  \image{figures/view_pred/#1/#2_input_views.png} &
  \image{figures/view_pred/#1/#2_query_views.png} &
  \image{figures/view_pred/#1/#2_pred_views.png} \\
}

\begin{figure}[h!]
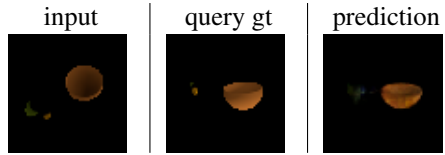

    \centering
    \begin{tabular}{c|c|c}
      input & query gt & prediction \\
      \generaterowsmall{singlein}{218200}
    \end{tabular}
    \vspace{0.2in}
    \caption{\textbf{View prediction} with a single image as input.}   
    \label{fig:singleinput}
\end{figure}

\newcommand{\imagew}[1]{
\begin{minipage}{.3\textwidth}
\includegraphics[width=\linewidth]{#1}
\end{minipage}
}

\newcommand{\generaterow}[2]{
  \imagew{figures/view_pred/#1/3d/#2_input_views.png} &
  \image{figures/view_pred/#1/3d/#2_query_views.png} &
  \image{figures/view_pred/#1/3d/#2_pred_views.png} &
  \image{figures/view_pred/#1/2d/#2_pred_views.png} \\
}

\newcommand{\generatefig}[3]{
\begin{figure*}[t!]
    \centering
    \begin{tabular}{c|c|c|c}
      Input V1,V2,V3 & query gt & \slamrnn & Tower \\
      \generaterow{#1}{000006}
      \generaterow{#1}{000007}
      \generaterow{#1}{000008}
      \generaterow{#1}{000009}
      \generaterow{#1}{000010}
      \generaterow{#1}{000011}
      \generaterow{#1}{000012}
      
      \generaterow{#1}{000013}
      \generaterow{#1}{000014}
      \generaterow{#1}{000015}
      \generaterow{#1}{000016}
      \generaterow{#1}{000017}
      
    \end{tabular}
    \caption{#2}
    \label{#3}
\end{figure*}
}


\generatefig{room}{View prediction results for the room scenes from ~\cite{Eslami1204}}{fig:room_results}

\generatefig{metzler}{View prediction results for the 7-segment shepard-metlzer dataset from ~\cite{Eslami1204}}{fig:metzler_results}

\generatefig{shapenet}{View prediction results for ShapeNet arrangement test scenes from ~\cite{activevision}}{fig:shapnet_results}

\generatefig{generalization}{View prediction results for 4-object scenes from ~\cite{activevision}}{fig:general_results}

\end{document}